\documentclass[sigconf]{acmart}

\makeatletter
\gdef\@copyrightpermission{
 \begin{minipage}{0.3\columnwidth}
  \href{https://creativecommons.org/licenses/by/4.0/}{\includegraphics[width=0.90\textwidth]{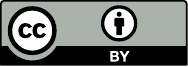}}
 \end{minipage}\hfill
 \begin{minipage}{0.7\columnwidth}
  \href{https://creativecommons.org/licenses/by/4.0/}{This work is licensed under a Creative Commons Attribution International 4.0 License.}
 \end{minipage}
 \vspace{5pt}
}
\makeatother

\copyrightyear{2024}
\acmYear{2024}
\setcopyright{rightsretained}
\acmConference[KDD '24] {Proceedings of the 30th ACM SIGKDD Conference on Knowledge Discovery and Data Mining }{August 25--29, 2024}{Barcelona, Spain.}
\acmBooktitle{Proceedings of the 30th ACM SIGKDD Conference on Knowledge Discovery and Data Mining (KDD '24), August 25--29, 2024, Barcelona, Spain}
\acmISBN{979-8-4007-0490-1/24/08}
\acmDOI{10.1145/3637528.3672045}

\usepackage{balance}

\usepackage{microtype}
\usepackage{graphicx}
\usepackage{subfigure}
\usepackage{booktabs} %
\usepackage{hyperref}

\usepackage{amsmath}
\usepackage{mathtools}
\usepackage{amsthm}

\usepackage[capitalize,noabbrev]{cleveref}

\usepackage{enumitem}
\usepackage{xspace}
\usepackage{autobreak}
\usepackage{multicol,multirow}
\usepackage{colortbl}
\usepackage{bm}

\newenvironment{itemize*}%
  {\begin{itemize}%
    \setlength{\itemsep}{2pt}%
    \setlength{\parskip}{2pt}}%
  {\end{itemize}}
\newenvironment{enumerate*}%
  {\begin{enumerate}%
    \setlength{\itemsep}{2pt}%
    \setlength{\parskip}{2pt}}%
  {\end{enumerate}}
\newenvironment{enumerate**}%
  {\begin{enumerate}%
    \setlength{\itemsep}{0pt}%
    \setlength{\parskip}{0pt}}%
  {\end{enumerate}}

\theoremstyle{plain}

\theoremstyle{definition}

\theoremstyle{remark}

\usepackage{algorithm}
\usepackage[noend]{algorithmic}

\newcommand{\method}{\texttt{RLHEX}\xspace}


\begin{document}

\title{Global Human-guided Counterfactual Explanations for Molecular Properties via Reinforcement Learning}

\author{Danqing Wang}
\authornote{Both authors contributed equally to this research.}
\affiliation{%
  \institution{Carnegie Mellon University}
  \department{Language Technologies Institute}
  \city{Pittsburgh}
  \country{USA}
}
\email{danqingw@andrew.cmu.edu}

\author{Antonis Antoniades}
\authornotemark[1]
\affiliation{
 \institution{University of California, Santa Barbara}
  \city{Santa Barbara}
  \country{USA}
}
\email{antonis@ucsb.edu}

\author{Kha-Dinh Luong}
\affiliation{%
 \institution{University of California, Santa Barbara}
  \city{Santa Barbara}
  \country{USA}
 }

\author{Edwin Zhang}
\affiliation{%
 \institution{Harvard University}
  \city{Cambridge}
  \country{USA}
 }
\affiliation{
 \institution{Founding}
 \city{Austin}
 \country{USA}
}

\author{Mert Kosan}
\authornote{Work done prior to joining Visa Inc.}
\affiliation{%
 \institution{University of California, Santa Barbara}
  \city{Santa Barbara}
  \country{USA}
 }

\author{Jiachen Li}
\affiliation{%
 \institution{University of California, Santa Barbara}
  \city{Santa Barbara}
  \country{USA}
 }

\author{Ambuj Singh}
\affiliation{%
 \institution{University of California, Santa Barbara}
  \city{Santa Barbara}
  \country{USA}
 }

 \author{William Yang Wang}
\affiliation{%
 \institution{University of California, Santa Barbara}
  \city{Santa Barbara}
  \country{USA}
 }

\author{Lei Li}
\affiliation{%
  \institution{Carnegie Mellon University}
  \department{Language Technologies Institute}
  \city{Pittsburgh}
  \country{USA}
}

\renewcommand{\shortauthors}{Danqing Wang, Antonis Antoniades et al.}

\begin{abstract}
Counterfactual explanations of Graph Neural Networks (GNNs) offer a powerful way to understand data that can naturally be represented by a graph structure. Furthermore, in many domains, it is highly desirable to derive data-driven global explanations or rules that can better explain the high-level properties of the models and data in question. However, evaluating global counterfactual explanations is hard in real-world datasets due to a lack of human-annotated ground truth, which limits their use in areas like molecular sciences. Additionally, the increasing scale of these datasets provides a challenge for random search-based methods.
In this paper, we develop a novel global explanation model \method for molecular property prediction. It aligns the counterfactual explanations with human-defined principles, making the explanations more interpretable and easy for experts to evaluate. \method includes a VAE-based graph generator to generate global explanations and an adapter to adjust the latent representation space to human-defined principles. Optimized by Proximal Policy Optimization (PPO), the global explanations produced by \method cover 4.12\% more input graphs and reduce the distance between the counterfactual explanation set and the input set by 0.47\% on average across three molecular datasets. \method provides a flexible framework to incorporate different human-designed principles into the counterfactual explanation generation process, aligning these explanations with domain expertise. The code and data are released at https://github.com/dqwang122/RLHEX.

\end{abstract}

\ccsdesc[500]{Computing methodologies~Artificial intelligence}
\ccsdesc[500]{Applied computing~Molecular sequence analysis}

\keywords{Graph Neural Network; Counterfactual Explanation; Reinforcement Learning}

\maketitle

\section{Introduction}
\label{sec:intro}
Graph Neural Networks (GNNs) have shown promise in fields such as cheminformatics and molecular sciences~\citep{gilmer2017neural,wieder2020compact}. A crucial application of GNNs in these fields is molecule property prediction~\citep{ross2022largescale,méndezlucio2022mole, doi:10.1073/pnas.2016239118}, where the task is to predict a molecule's properties based on its structural or functional groups. This is essential for various scientific research aspects, including drug discovery~\citep{gomez2018automatic}, environmental monitoring~\citep{nguyen2020environmental}, and materials engineering~\citep{butler2018machine}.

However, the intricate complexity of Graph Neural Networks (GNNs) poses challenges in fully leveraging the rich information embedded within the structural properties and feature representations of nodes and edges~\citep{zhang2020deep,zhou2020graph,wong2023discovery}. Moreover, it is challenging to interpret and understand the underlying rationale behind GNNs' prediction due to non-transparency.
This complexity underlines a growing need to understand GNN predictions, particularly with counterfactual (CF) explanations, which present the conditions that need to change to alter the model's decisions~\citep{abrate2021counterfactual,lucic2022cf}. They can highlight the influential sub-graph affecting an individual graph's prediction (local explanations) ~\citep{lucic2022cf,tan2022learning,ma2022clear}, or a collection of factors affecting the whole dataset that outlines the model's learned principles (global explanations) ~\citep{yuan2020xgnn,gcfexplainer2023}.

Local explanations are known to be vulnerable to a small noise in the input graph~\citep{bajaj2021robust} and often fail to generalize to new graphs~\citep{ma2022clear}. Moreover, without ground-truth labels for local explanations in real-world datasets, it is challenging to evaluate the performance on large datasets. For example, \citet{gnome} predicts 2.2 million molecules as crystals, making it feasible to manually check the explanation for each prediction without the ground-truth labels. This underscores a significant need for more comprehensive and global explanations. 

\begin{figure*}[th]
    \centering
    \includegraphics[width=0.95\linewidth]{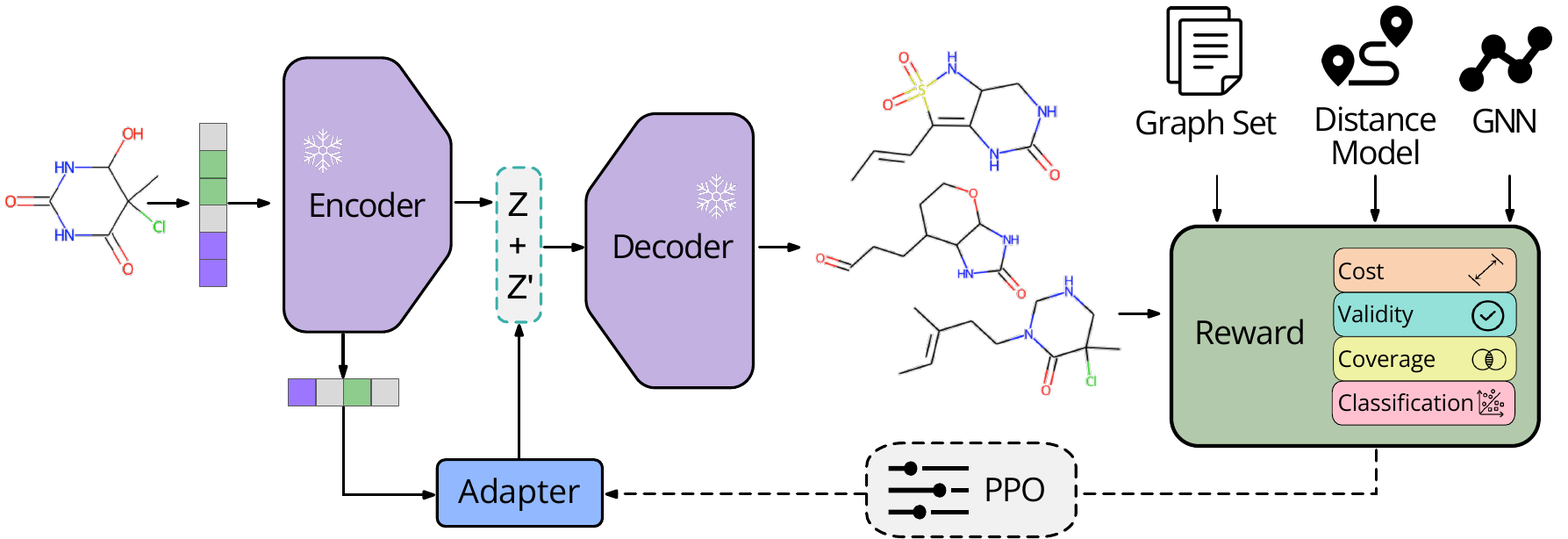}
    \caption{\method has three main parts - the VAE-based generation model, the adapter, and the reward module. The reward module contains several reward functions based on principles designed by humans, which make the generated explanations easier for domain experts to interpret. The adapter modifies the latent representation $z$ by adding the delta $z'$, which is optimized using PPO to align with the principles designed by humans. The \method model uses the molecule to be explained as the input and creates the CF explanations from the modified latent representation $z + z'$.}
    \label{fig:overview}
\end{figure*}

Global explainers offer a small set of explanations for the whole input set's behavior. They are more resistant to changes and easier for humans to assess. However, finding global explanations that adequately explain the GNN decision is challenging. \citet{gcfexplainer2023} shows it is an NP-hard problem to find the best CF explanation set that can cover as many input graphs as possible regarding the size limit of the set. Additionally, previous CF explainers often use individual nodes, edges, or features as explanations, which are hard for domain experts to verify. For example, ~\citet{wu2023chemistry} states that previous explanations for predicting molecule properties do not match chemists' intuition as these features are not chemically meaningful. Chemists rely on bioisosteres and functional groups to identify the properties of a molecule. Resolving this issue could significantly improve the work of chemists, allowing them to make well-informed decisions based on interpreted results from molecule property prediction.

In response to these challenges, we introduce a global CF explanation model to aid domain experts in understanding the high-level behavior of GNN predictors, which is called \textbf{R}einforcement \textbf{L}earning via \textbf{H}uman-guided \textbf{EX}planations (\method). \method first identifies several principles for the ideal global CF explanation to align with the preference of domain experts. For example, the global CF explanations should be chemically valid and cover as many input molecules as possible. \method then employs a Variational Autoencoder (VAE)-based molecule generation model to generate global CF explanations and introduces an adapter to align it with the human-designed principles. Specifically, the adapter learns the policy for aligning the initial molecule latent space with the desired CF explanations' space by leveraging Proximal Policy Optimization (PPO).
By sampling from the aligned latent space, the CF explanations generated by \method outperform other strong baselines on three molecular datasets. In summary, our contributions are:

\begin{itemize}[nosep]
    \item We propose a novel counterfactual explanation model for GNNs to generate global CF explanations that align with human-designed principles. These concise explanations offer domain experts a better insight into GNNs' prediction rationale. 
    \item By optimizing the latent space via PPO, \method can sample diverse explanations satisfying desirable and diverse principles.
    \item Experimental results show that \method can achieve the best performance on three real-world molecule datasets: AIDS~\citep{riesen2008iam}, Mutagenicity~\citep{kazius2005derivation} and Dipole~\citep{pereira2018machine} in terms of several evaluation metrics.
\end{itemize}

\section{Related work}
\label{sec:related}
\paragraph{Counterfactual Explanations of GNNs}
Counterfactual reasoning presents a necessary condition that would, if not met, alter the prediction~\cite{prado2023survey, kosan2024gnnxbench}. Studies aimed at providing counterfactual explanations for GNNs can be divided into two categories: \textit{local} and \textit{global} explainers. Local explainers select sub-structures from a given graph that contribute to its GNN's prediction~\citep{lucic2022cf,tan2022learning,numeroso2021meg}, whereas global explainers produce new graphs to illustrate the model behavior across a set of graphs~\citep{yuan2020xgnn,gcfexplainer2023}. There are two typical ways to generate counterfactual explanations. One is to perturb nodes and edges of the input graph to get a different prediction. For example, ~\citet{lucic2022cf} and ~\citet{tan2022learning} learn a mask matrix to select the sub-graph or feature from the input graph, while ~\citet{gcfexplainer2023} and ~\citep{10.1145/3580305.3599289} explore different graph edits. The other one is to model it as a generative task. For example, ~\citet{yuan2020xgnn}, ~\citet{numeroso2021meg} and ~\citet{ma2022clear} generates the key pattern for counterfactual explanations directly. However, the explanations created by previous methods are difficult to evaluate without the ground-truth labels. In this paper, we align a graph generative model with human-designed principles to make the explanations more human-friendly.

\paragraph{Graph-based Molecule Generation}
Graph neural networks are widely used in 2D molecule tasks~\citep{gilmer2017neural,wieder2020compact}. A molecule can be graphically represented with atoms as vertices and chemical bonds as edges. The atom-based generation methods take the atom as the basic generation units~\citep{li2018learning,you2018graph}, while the fragment-based methods build their vocabulary based on the chemical substructure~\cite{jin2018junction,liu2017break,kong2022molecule}. The fragment-based generation is more likely to produce meaningful molecules with chemically desirable properties, which are reflected in their substructure~\citep{wu2023chemistry,geng2023novo}. Additionally, it can make the edit-based sampling more effective and efficient~\cite{xiemars}. In this paper, we use a fragment-based generative model to ensure that the generated global explanations are valid molecules, making them understandable for domain experts.

\paragraph{Align Explainability with Humans}
Aligning models to make them helpful and friendly to humans has gained increasing attention recently, especially in large language models~\cite{ouyang2022training,bai2022training}. The alignment can be formulated in the reinforcement learning framework to optimize the model policy towards the reward functions based on human values~\citep{bai2022constitutional} or principles~\cite{sun2023salmon,sun2023principledriven}. ~\cite{lai2023selective} explores the selective explanations based on what aligns with the recipient's preferences. ~\citet{xu2023instructscore} investigates explainable metrics and aligns the explanation of the mistakes with humans by the failure mode summarized from human feedback. However, few studies have investigated how to align the global interpretation of GNNs with the preferences of domain experts.

\section{Preliminaries}
\label{sec:background}
\subsection{Molecular Property Prediction}
A molecule can be intuitively represented as a graph $G=(V,E)$, where $V \in \mathbb{R}^{|V| \times d_{v}}$ is a set of nodes corresponding to atoms and $E \in \mathbb{R}^{|V| \times |V| \times d_{e}} $ is a set of edges corresponding to chemical bonds. $|V|$ is the number of atoms. $d_{v}$ and $d_{e}$ are the feature dimensions of the atom and the bonds respectively.
The functional groups of the molecule are often denoted by a subgraph $S=(V_s, E_s)$ of the graph $G$, which satisfies $V_s \in V$ and $E_s \in E$. 

The molecular property prediction can be modeled as a binary graph classification task. It aims to predict whether the input molecule has a certain chemical property, such as whether the molecule is active against AIDS. Given a set of $n$ molecular graphs $\mathbb{G}=\{G_0, G_2, \cdots, G_n\}$ and the ground-truth labels $\{y_0, y_1, \cdots, y_n\}$ where $y_i \in \{0, 1\}$, the GNN classifier $f_{\phi}(\cdot)$ is trained to predict the estimated label $\hat{y_i}=f_{\phi}(G_i)$ for each input graph $G_i$.

Typically, GNN learns the representation of each node $h_v \in \mathbb{R}^d$ by aggregating the information of its neighbors $N(v)$~\citep{gilmer2017neural}. $d$ is the dimension of the hidden state. By identifying different aggregation functions $\mathrm{M}_{\phi}(\cdot)$ and node update functions $\mathrm{U}_{\phi}(\cdot)$, the update mechanism in each layer $l$ can be denoted as:
\begin{align}
    \label{eqn:message} m_v^{l} &= \sum_{w \in N(v)} \mathrm{M}_{\phi}(h_v^{l-1}, h_w^{l-1}, e_{vw}) \\
    \label{eqn:update} h_v^{l} &= \mathrm{U}_{\phi}(h_v^{l-1}, m_v^l).
\end{align}
Here $e_{vw}$ is the feature vector of the edge between node $v$ and $w$. Finally, a pooling layer is added on top of the last hidden layer $L$ to get the representation of the whole graph:
\begin{align}
    \label{eqn:readout}
    h_{G} = \text{Pooling}(\{h_{v}^{L} | v \in V\}),
\end{align}
and a classification head is used to predict binary label $\hat{y}$ for the given chemical property.

\subsection{GNN Counterfactual Explanation}

The explanation of the GNN classifier is to analyze and interpret how it makes predictions. A local counterfactual explanation (CF explanation) of the GNN classifier $f_{\phi}(\cdot)$ on the input molecular graph $G$ is defined as an instance $C$ that $f_{\phi}(C) \neq f_{\phi}(G)$. Here, $C$ can be the sub-graph of the original $G$ or another graph similar to $G$. The optimal CF explanation $C^{*}$ is one that minimizes the distance between $G$ and the CF explanation~\citep{lucic2022cf}. Ideally, the optimal CF explanation should be very close to the input graph and have a different prediction. It reveals the minimal perturbation the classifier needs to change its decision.

For the global CF explanation, it aims to provide a set of $K$ instances $\mathbb{C}=\{C_1, C_2, \cdots, C_K\}$ that can explain the global behavior of the classifier. Different from the local CF explanations which is specifically designed for each input molecule, the limited size of the global CF explanation set makes it easier for domain experts to check the classifier behavior on large molecule datasets. For example, given an undesirable molecule property $y=0$ and a set of molecules $\{G_i | f_{\phi}(G_i)=0, G_i \in \mathbb{G} \}$, the global CF explanation set is a set of $K$ instances $\{C_k | f_{\phi}(C_k) \neq 0, k \in \{0, 1, \cdots, K\}\}$. \citet{gcfexplainer2023} proposes that the global CF set should have a high \textit{coverage}, a low \textit{cost} and a small \textit{size}.

\subsection{VAE-based Graph Generation}
Given the input graph $G=(V,E)$, the variational auto-encoders first embed the graph into continuous latent representation $\bm{z} \in \mathbb{R}^{d_z}$ by the encoder $p_{\varphi}(\bm{z} | G)$. $d_{z}$ is the dimension of the latent representation. The graph decoder then outputs the graph from the sampled point in the latent space $q_{\psi}(\hat{G} | \bm{z})$~\citep{simonovsky2018graphvae}. The model is trained by minimizing the training objective:
\begin{align}
\label{eqn:elbo}
L = - E_{q_{\psi}(\bm{z}|G)}[\log p_{\varphi}(G|\bm{z})] + \text{KL}(q_{\psi}(\bm{z}|G) || p(\bm{z})).
\end{align}
Here, $p(z)$ is the prior distribution $\mathcal{N}(0,1)$. It maximizes the likelihood of the input graph $G$ and regularizes the latent space with the KL divergence.

\textbf{Encoder} The graph neural network is often used as the encoder to map the input graph into the latent representation $\bm{z}$. The hidden representation of the graph $h_G$ is obtained by Eqn \ref{eqn:readout}. It is then mapped to the $\mu_{G}$ and log variance $\sigma_{G}$ of variational posterior approximation $q_{\psi}(\bm{z}|G)$ for the reparameterization~\citep{kingma2013auto}. The latent representation is sampled from $\mathcal{N}(\mu_{G}, \sigma_{G}|G)$.

\textbf{Decoder} Based on how the graph is generated, there are two typical types of graph decoder. One is to pre-define the number of nodes $|V|$ and then predict the node feature matrix $\hat{V}$ and edge matrix $\hat{E}$. The other is to autoregressively generate nodes $P(v_i | v_{<i}, \bm{z}), i \in \{1,\cdots, |V|\}$ and then predict the edge between the nodes $P(e_{v,w} | \bm{z})$.

\section{Methodology}
\label{sec:method}
In this section, we propose a novel global explanation framework \method to align global CF explanations with human-designed principles. We first discuss several desirable properties for the optimal global CF explanation of molecules. Then we introduce the backbone framework to generate CF candidates and use Proximal Policy Optimization (PPO) to align the candidates with these human principles. 

\subsection{Principle of Global Molecule Explanation}
\label{sec:principle}
Inspired by previous studies, we investigate three principles to guide the generation to make the global explanation more understandable and interpretable to domain experts such as chemists. 

\begin{enumerate}[nosep]
    \item \textit{The generated explanations should be counterfactual to the input molecules}~\cite{prado2023survey}.
    \item \textit{The generated explanations should be valid molecule.}~\citep{wu2023chemistry}.
    \item \textit{The explanation set should be small enough for an expert to manually evaluate while covering as many input molecules as possible}~\citep{gcfexplainer2023}.
\end{enumerate}

The first one is the basic principle for CF explanation, requiring the counterfactual explanation $C$ to have a different prediction with the input molecule $G$: $f_{\phi}(C) \neq f_{\phi}(G)$.
The second one ensures that the generated explanations are chemically meaningful structures to chemists. This interpretability is more compatible with the domain knowledge and easy for the chemists to understand. For example, the generated graphs should not violate the implicit valence and ring information.
The last one is derived from the definition of the optimal local CF explanation. The CF explanation set should be similar to the input molecule so that it can reveal the necessary features the GNN predictor relies on to change their prediction. The size of the explanation should also be small to ensure it is durable for domain experts to check. Here we follow \citet{gcfexplainer2023} to introduce three metrics to formally define the requirement: \textbf{cov}, \textbf{cost}, and \textbf{size}. The size is denoted as $|\mathbb{C}|$.

Coverage is a measure of the proportion of input graphs $G \in \mathbb{G}$ can be covered by explanations in $\mathbb{C}$ under a given distance threshold $\delta$:
\begin{align}
    \text{\textbf{cov}} (\mathbb{C}) = \frac{|\{G \in \mathbb{G} | \min_{C \in \mathbb{C}} {d(G, C) \leq \delta}\}|}{|\mathbb{G}|},
\end{align}
where $d(G, C)$ is the function to calculate the similarity between two graphs, and $|\mathbb{G}|$ indicates the size of the set $\mathbb{G}$. The cost is the mean distance between the input graph set $\mathbb{G}$ and the explanation set $\mathbb{C}$:
\begin{align}
    \text{\textbf{ cost}} (\mathbb{C}) = \frac{1}{|\mathbb{G}|} \sum_{i=1}^{|\mathbb{G}|} \min_{C \in \mathbb{C}} d(G, C).
\end{align}

\subsection{Adapter-enhanced Molecule Generator}

To align these human-designed principles with CF generation, we propose Reinforcement Learning via Human-guided EXplanations~\method, a flexible CF generation framework for molecules. It formulates the search for an optimal global counterfactual explanation as a graph set generation task. Given the input graph set $\mathbb{G}$ and the GNN classifier $f_{\phi}(\cdot)$, \method generates a set of graphs $\mathbb{C}$ as its CF explanations.

As shown in Figure \ref{fig:overview}, \method includes three modules: a VAE-based generation model, an adapter, and a reward module.

\begin{itemize}[nosep]
    \item \textbf{VAE-based Generation Model}. It aims to sample diverse valid molecules as CF candidates.
    \item \textbf{Adapter} module. It is a parameterized policy $\pi_{\theta}$ to steer the generator into producing explanations that meet the human-designed principle.
    \item \textbf{Reward Module}. It provides the reward signal based on the human-designed principles to guide the generation. Both heuristic-based and parameterized criteria can be flexibly integrated to meet specific experimental needs or to customize explanations as required.
\end{itemize}

\subsubsection{VAE-based Generation Model}
We backbone our model with a fragment-based molecule generation model Principal Subgraph VAE (PSVAE) \cite{kong2022molecule} $M_{\psi}$. It first mines principal subgraphs from the molecule datasets and then generates new molecules based on the subgraphs. Essentially, principal subgraphs are frequent and large fragments. The sub-graph-based generation is more interpretable and chemically meaningful, resulting in valid molecules.

PSVAE decomposes one molecule into a set of unordered non-overlapped principal sub-graphs $[F_0, \cdots, F_n]$. $n$ is the number of subgraphs.
To generate a new molecule, PSVAE autoregressively predicts chemical sub-graphs $P(F_i | F_{<i}, \bm{z})$. It then non-autoregressively predicts the inter-subgraph edges $e_{vw}$ via GNN, where $v$ and $w$ are nodes from different sub-graphs. Therefore, the likelihood of the generated molecule can be formulated as:
\begin{align}
    \label{eqn:psvae}
    \sum_{i=0}^{n} \log P(F_i | F_{<i}, \bm{z}) + \sum_{v \in F_i, w \in F_j, i \neq j} \log P(e_{vw}|\bm{z}).
\end{align}
By replacing the first term of Eqn \ref{eqn:elbo} with Eqn \ref{eqn:psvae}, we obtain the training objective of PSVAE. We use the pre-trained checkpoint of PSVAE and freeze the parameters in the encoder and decoder.

\subsubsection{Latent Distribution Adaptor}

To steer the molecule generation model to create desired CF explanations, we add a lightweight adaptor as our parameterized policy $\pi_{\theta}$ to adjust the latent distribution. The adaptor can either be initialized as a copy of the initial PSVAE, denoted as $M_{\psi}$, or as a randomly initialized model, in our case a lightweight transformer encoder~\citep{vaswani2017attention}. It takes the graph hidden state $h_{G}$ as the input and maps it to a shift on the mean of the latent distribution $\mu'_{G} = \pi_{\theta}(h_{G})$. The new latent representation $\bm{z'}$ is sample from the distribution $\mathcal{N}(\mu_{G} + \mu'_{G}, \sigma_{G})$. The decoder takes $z'$ as the input to generate $q_{\psi}(G | \bm{z'})$.  The complete generative process, starting from the input molecule $G_t$ and ending at the explanation molecule $C_t$ can be formalized as follows:
\begin{align}
    \label{eqn:encode} \mu_{G_t}, \sigma_{G_t} &= M_{\psi}^{(E)}(G_t) \\
    \label{eqn:shift} \mu'_G &= \mu_G + \pi_{\theta}(h_{G_{t}}) \\
    \label{eqn:sample} z' & \sim \mathcal{N}(\mu_{G_t} + \mu'_{G_t}, \sigma_{G_t}) \\
    \label{eqn:decode} C_t &= M_{\psi}^{(D)}(z'),
\end{align}
where and $M_{\psi}^{(E)}$, $M_{\psi}^{(D)}$ stand for the respective encoder and decoder of the PSVAE. This framework allows for the sequential generation of molecules, with each step informed by the previous state, thus enabling a guided exploration of the molecular space that is coherent with the desired properties encoded by the policy $\pi_{\theta}$.

\subsubsection{Principle Modeling}
\label{sec:reward}
We design our reward module based on the principles in Section \ref{sec:principle}. 
For (1), we take the prediction probability of the opposite class as the reward. For example, if the input molecules are predicted as negative $p\left(f_{\phi}(G)=0\right)$, we take the probability of $p\left(f_{\phi}(C)=1\right)$ as the reward. We use $p_{\phi}(C)$ for short.
For validity in (2), we use the Rdkit library~\footnote{https://www.rdkit.org/} to build an indicator function $\mathbb{I}(\cdot)$ as the reward signal~\footnote{\textit{rdkit.Chem.detectChemistryProblems} is used to check molecule validity. It inspects molecular properties such as atom valence. Experts can easily add more constraints such as QED~\citep{bickerton2012quantifying} by replacing this function.}. If the molecule is valid, $\mathbb{I}(C)=1$, otherwise, $\mathbb{I}(C)=0$. 

The objective of (3) can be written as:
\begin{align}
    \max_{\mathbb{C}}\text{\textbf{cov}} (\mathbb{C}) \quad s.t. |\mathbb{C}|=k.
\end{align}
Here, the size of $\mathbb{C}$ is limited by k, and the cost is constrained based on the threshold $\delta$ in coverage.  
In practice, we first maximize the local reward for each input molecule $G$ and get a set of CF explanation $\mathbb{C}$. Then we greedily select top-$k$ explanations from the candidate set as $\mathbb{C}_{k}$. The local reward $s(C)$ for each candidate $C$ is defined as:
\begin{align}
    \label{eqn:reward_ind}
    \textbf{score}(C) = \mathbb{I}(C) * [\alpha p_{\phi}(C) + \beta \text{\textbf{cov}}(C)]
\end{align}
Here $\alpha, \beta$ are the coefficients of the prediction probability and the local coverage. If we use $R(\mathbb{C})$ to indicate the sum of local reward on the set $\mathbb{C}$, the optimal global score of the CF explanation set with size $k$ can be written as:
\begin{align}
    \label{eqn:reward}
    R^{*}(\mathbb{C}_k) = \max_{\mathbb{C}_k} \sum_{C \in \mathbb{C}_k} \textbf{score}(C), \quad s.t. |\mathbb{C}_k|=k.
\end{align}

\subsection{Tailor Latent Distribution via PPO}
We formulate the alignment to the human-designed principles as a Markov decision process (MDP) $\mathcal{M}$ = ($\mathcal{S}, \mathcal{A}, \mathcal{R}, p$), with state space $\mathcal{S}$,
action space $\mathcal{A}$, reward function $\mathcal{R}$, and transition probability matrix $p$. 
The state $s \in \mathcal{S}$ is the CF candidates, which are all molecules. The action is the modification of the latent distribution $\mu'_{G} \in \mathcal{A} \subseteq \mathbb{R}^{d_z} $. $d_z$ is the dimension of the latent space. $p(\cdot | s, a): \mathcal{S} \times \mathcal{A} \to \mathcal{S}$ indicates the probability of CF candidates based on the new latent distribution.
The reward $r \in \mathcal{R}$ is the local reward function defined in Eqn \ref{eqn:reward_ind}.

At each time step $t$, \method employs the VAE encoder to get the embedding of the input molecule $G$, which is the state $s_t$. It then uses $\pi_{\theta}(s_t, a_t)$ to sample the mean shift of the latent distribution $\mu'(G)$. The reward $r(s_t, a_t)$ is based on the scores from the human-designed principle (Eqn \ref{eqn:reward_ind}). Our goal is to learn a policy $a \sim \pi_{\theta}(s)$ that can find the optimal latent distribution for the CF explanations.

\begin{algorithm}[t]
    \caption{\method Inference}
    \label{alg:inference}
    {\small
    \begin{algorithmic}[1]
    \STATE The input molecule set $\mathbb{G}$, the encoder $M_{\psi}^{(E)}$ and the decoder $M_{\psi}^{(D)}$, the policy $\pi_{\theta}$, the CF set $\mathbb{C}$
    \STATE $\mathbb{C} \gets \emptyset$
    \FOR{$G \in \mathbb{G}$}
    \FOR{$t \in 1:T$}
    \STATE Map $G$ to the latent distribution by $M_{\psi}^{(E)}$ and $\pi_{\theta}$ via Eqn \ref{eqn:encode} and \ref{eqn:shift}
    \STATE Sample $z' \gets \mathcal{N}(\mu_{G} + \mu'_{G}, \sigma_{G)}$ via Eqn \ref{eqn:sample}
    \STATE Get one candidate $C$ by $M_{\psi}^{(D)}$ via Eqn \ref{eqn:decode} 
    \STATE $G \gets C$
    \ENDFOR
    \STATE $\mathbb{C} \gets \mathbb{C} + \{C\}$
    \ENDFOR
    
    \FOR{$i \in 1:k$}
    \STATE $C \gets \text{argmax}_{C \in \mathbb{C}} \Delta r(C ; \mathbb{C}_{k-1})$
    \STATE $\mathbb{C}_k \gets \mathbb{C}_k + \{C\}$
    \ENDFOR
    \end{algorithmic}}
\end{algorithm}

We leverage Proximal Policy Optimization (PPO) \cite{schulman2017proximal} to generate molecules that satisfy different principles. It has been extensively used to steer models into producing human-desired outputs in language models, using RLHF (Reinforcement Learning from Human Feedback) \cite{DBLP:journals/corr/abs-1909-08593, ouyang2022training, christiano2023deep}, and we take inspiration from these methods to build a more flexible system within which many different forms of human-guided principles can be used to optimize our explanation model.

PPO operates by optimizing an objective that balances exploration and exploitation of the current policy $\pi_{\text{old}}$, to gain more reward and explore new policies $\pi_{\theta}$. This balance is achieved through clipped probability ratios, ensuring that updates do not deviate too far from the current policy. The standard PPO objective is the following:

\begin{align}
\label{eqn:ppo}
L(\theta) = \mathop{\mathbb{E}}_{\substack{s_t \sim p(\cdot|s_{t-1}, a_{t-1}) \\ a_t \sim \pi_{old}(\cdot | s_{t-1})}} \Bigg[ &\min \Bigg( \frac{\pi_{\theta}(a_t|s_t)}{\pi_{\text{old}}(a_t|s_t)} \hat{A}(s_t, a_t), \\
&\text{clip}\left( \frac{\pi_{\theta}(a_t|s_t)}{\pi_{\text{old}}(a_t|s_t)}, 1 - \epsilon, 1 + \epsilon \right) \hat{A}(s_t, a_t) \Bigg) \Bigg], \nonumber
\end{align}
where $\epsilon$ is a hyperparameter that defines the clipping range, and $\mathbb{E}_{s_t, a_t}$ represents the expectation for an on-policy batch sample. We train an additional critic model $V(s)$ to estimate the actual reward of the current state and action $Q(s_t, a_t)$ during the training. Here $Q(s_t, a_t)$ is the expected local rewards of the CF candidates sampled from the new latent distribution: $Q(s_t, a_t) = \mathbb{E}_{C \sim p(\cdot|s_t,a_t)}[\textbf{score}(C)]$. $\textbf{score}(C)$ is from Eqn \ref{eqn:reward_ind}.
$\hat{A}(s_t, a_t)$ denotes the advantage function, which is defined as $A(s, a) = Q(s, a) - V(s)$.

In our case, the property of a constrained policy update is essential to our goal of keeping newly generated molecules close in distribution to the original molecule distribution to ensure their validity.

\subsection{Greedy Selection}
For each molecule $G$ in $\mathbb{G}$, we apply \method to optimize the local reward $\textbf{score}(C)$ in Eqn \ref{eqn:reward_ind} and get the CF candidate set $\mathbb{C}$. To get an optimal CF candidate set $\mathbb{C}_k$ with size $k$, we greedily choose the top-$k$ candidates from $\mathbb{C}$. Start from an empty set $\mathbb{C}_0$, we add the candidate with the maximum gain, which is defined as:
\begin{align}
    \textbf{Gain}(C; \mathbb{C}) = R(\mathbb{C} + C) - R(\mathbb{C}).
\end{align}
The detailed phase is described in Alg. \ref{alg:inference}.

\section{Experiment}
\label{sec:exper}
\subsection{Datasets}

We focus on molecule property prediction and conduct our experiments on three real-world molecule datasets: AIDS~\citep{riesen2008iam}, Mutagenicity~\citep{kazius2005derivation} and Dipole~\citep{pereira2018machine}. AIDS and Mutagenicity have been used in previous works~\citep{gcfexplainer2023,wang2022reinforced}, however, qualitative evaluation of the counterfactuals generated for these tasks is challenging without domain expertise. For that reason, we seek to formulate a task in which the chemical characteristics can be quickly evaluated by observing the structure of the generated graphs. AIDS is a binary dataset where the label=0 indicates the molecule is active against AIDS. The activity is the desirable attribution for molecules, so we flip the label to make the negative class correspond to the undesirable property.
The mutagenicity dataset classifies molecules by whether they are mutagenetic and labels the mutagenetic with label 0.
Dipole is a binary classification dataset we curated from a subset of molecules reported by \citet{pereira2018machine}, in which the dipole moment of each molecule is recorded. In particular, we extract a subset of the most polar molecules to form the positive class and a subset of the least polar molecules to form the negative class. Polarity is a comparably simpler chemical property to assess from only the molecular structure.
Following previous work, we keep atom types that appear at least 50 times in the dataset, resulting in 9 common atoms in AIDS and Dipole and 10 in Mutagenicity.~\footnote{We further remove molecules with atom \textit{Na} because it is not a common atom in ZINC250K pre-trained dataset~\citep{irwin2012zinc}.} For the graph dataset AIDS and Mutagenicity, we convert the graph representation to SMILES and remove the duplicated instances. 

We randomly split the dataset by 0.8:0.1:0.1 for training, validation and testing.
We follow \citet{gcfexplainer2023} to train separate GNN-based predictors $f_{\phi}(\cdot)$ on these datasets. We use 3 convolution layers as the aggregation function and add the message to the previous node representation for update. One max pooling layer is used to get the graph representation $h_G$ from the node representation and a full-connected layer is added on top of it for classification. 
The model is trained with the Adam optimizer~\citep{DBLP:journals/corr/KingmaB14} and a learning rate of 0.001 for 1000 epochs. Detailed information is listed in Table \ref{tab:dataset}. 

\begin{table}[htbp]
  \centering
  \caption{Dataset Statistic. Here \# indicates the number size. We train the GNN classifiers to predict the molecule property. We ignore atoms that appear less than 50 times in the dataset and filter the duplicated molecules by SMILES representation.}
    \begin{tabular}{lccc}
    \toprule
          & AIDS  & Mutagenicity & Dipole \\
    \midrule
    \#Graphs &  1562 &  3461 & 3539\\
    \#Nodes per graph & 15.73 & 30.34 & 9.16\\
    \#Edges per graph & 16.32 & 30.80 & 18.53\\
    \#Atom Type & 9 & 10 & 9 \\
    \#GNN Accuracy & 97.81\% & 80.00\% & 89.37\% \\
    \bottomrule
    \end{tabular}%
  \label{tab:dataset}%
\end{table}%

\subsection{Baselines}
Since most CF explanation methods for GNNs focus on local explanations, which are not directly comparable, we present the state-of-the-art global explanation method \textbf{GCFExplainer}~\cite{gcfexplainer2023} and elaborate two sampling-based generative baselines.
\textbf{GCFExplainer} functions employ vertex-reinforced random walks on an edit map of graphs and a greedy summary to deliver high-coverage, low-cost candidate sets for input graphs. The maximum steps of the random walk is set to 6000. 
Two non-RL generative baselines are tested: \textbf{PSVAE} and \textbf{PSVAE-SA}. PSVAE iteratively encodes the input molecule to the latent space and samples from the latent space to generate candidates. We terminate the sampling process when its generation has a different prediction from the input molecule, or when it arrives at the maximum iteration.
PSVAE-SA applies simulated annealing to optimize the sampling process toward the reward function. For each iteration, it samples from the latent representation of the input molecule and accepts the new generation based on the Metropolis criterion: 
\begin{align}
    A_t = \min(1, e^{\frac{\text{\textbf{score}}_t - \text{\textbf{score}}_{t-1}}{T}}),
\end{align}
Here, $\text{\textbf{score}}_t$ represents the local reward at step $t$ defined in Eqn \ref{eqn:reward_ind}, while $T$ is temperature, initially set at 0.1 and halved every 10 steps. We follow the standard settings of other hyperparameters in original papers. We check the validity of the generated candidate explanations and only keep valid molecules with opposite predictions for the final candidate set.

\begin{figure*}
    \centering
    \includegraphics[width=0.9\linewidth]{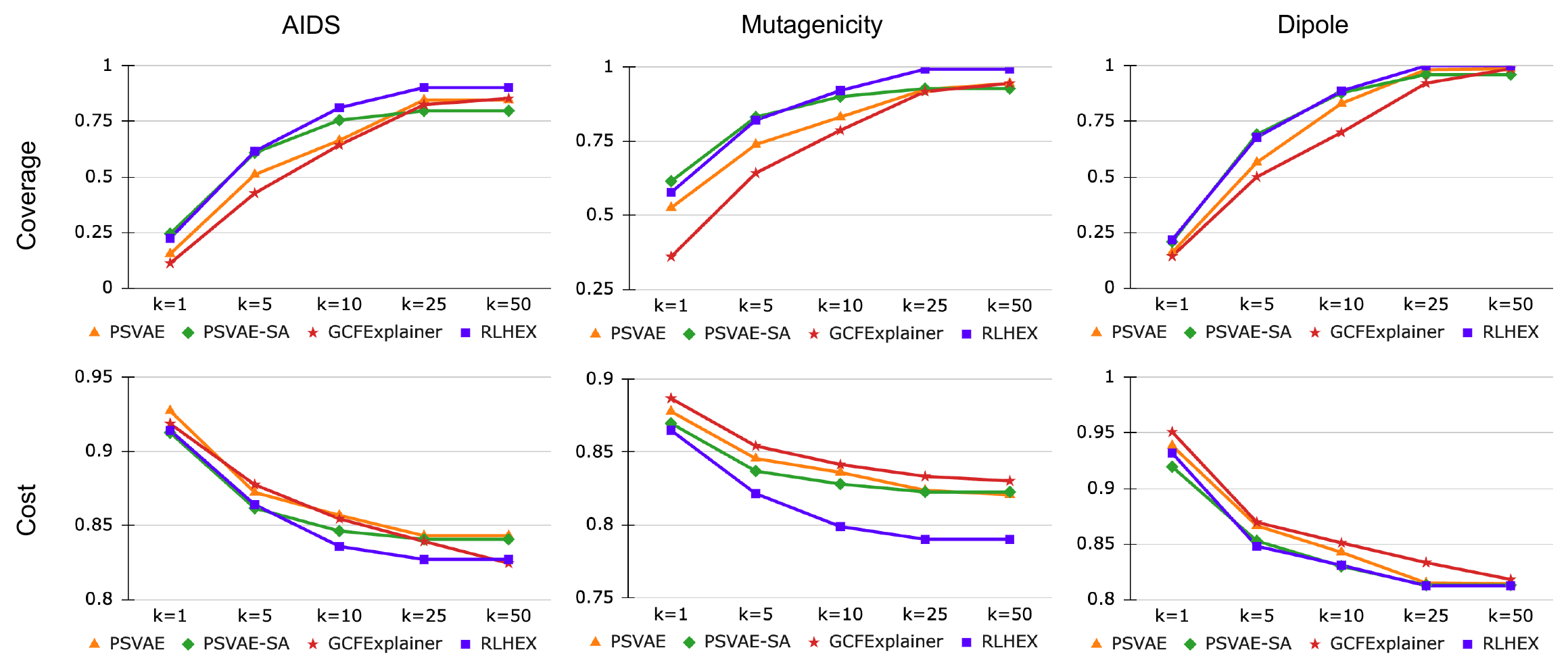}
    \caption{Coverage and Cost with different size $k$ of explanation set. \method generally outperforms the baselines on coverage and cost. We use iteration $i = 20$ for generation.}
    \label{fig:topk}
\end{figure*}

\subsection{Implementation and Evaluation}
We initialize the encoder and decoder of \method based on PSVAE~\cite{kong2022molecule}. We use the released checkpoint trained on ZINC250K~\citep{irwin2012zinc} and freeze the parameters. Note that our method does not limit us to a specific architecture, allowing for the use of other models for molecule generation purposes. We set the dimension of the latent representation space as 56 and the hidden size of the adaptor to 400 for both the policy model and the critic model. We use the Adam optimizer for training and set the learning rate to 1e-5. We use linear warmup and decay the learning rate to 1/10 of the maximum. 
To facilitate good explorations vs. exploitation strategy, we employ Upper Confidence Bound sampling \citet{wang2022thompson}, which preferably samples input molecules according to the mean and variance of the scores they yield over episodes.  

We follow previous studies to use Tanimoto similarity to calculate the distance between molecules~\citep{numeroso2021meg,wong2023discovery}. It is a commonly used similarity function to compare chemical structures based on fingerprints. We set the distance threshold $\delta$ as 0.87 based on the dataset distribution. 

We evaluate the model performance based on the coverage and cost described in Section \ref{sec:reward}. We set the coefficient in Eqn \ref{eqn:reward_ind} to $\alpha = 1, \beta=10$ to balance the scale of the prediction probability and the individual coverage. 
PSVAE-based baselines maintain a beam size of 10 and a temperature of 1 for decoding. After getting the counterfactual candidate set, we use the greedy algorithm to select top-$k$ counterfactual explanations as described in Alg. \ref{alg:inference}. The main results come from sampling $T=20$ iterations per input molecule. We use $k=10$ for the main experiment.

\begin{table}
\centering
\caption{Coverage measures the percentage of the input molecules covered by the top-10 counterfactual explanations. \method achieves the highest coverage on three datasets.}
\label{tab:coverage}
\begin{tabular}{lccc}
\toprule
Coverage $\uparrow$ & AIDS & Mutagenicity & Dipole \\
\midrule
GCFExplainer & 0.647$\pm$0.016 & 0.806$\pm$0.043 & 0.762$\pm$0.020 \\
PSVAE & 0.682$\pm$0.030 & 0.850$\pm$0.023 & 0.836$\pm$0.018 \\
PSVAE-SA & 0.761$\pm$0.024 & 0.894$\pm$0.011 & 0.872$\pm$0.010 \\
\method & \textbf{0.822}$\pm$0.021 & \textbf{0.909}$\pm$0.024 & \textbf{0.896}$\pm$0.017 \\
\bottomrule
\end{tabular}
\end{table}
\begin{table}
\centering
\caption{Cost indicates the minimum distance between the input set and the top-10 explanation set. \method creates counterfactual explanations that are more similar to the input graphs but have different GNN predictions.}
\label{tab:cost}
\begin{tabular}{lccc}
\toprule
Cost $\downarrow$ & AIDS &Mutagenicity &Dipole \\
\midrule
GCFExplainer & 0.853$\pm$0.008 & 0.917$\pm$0.046 & 0.964$\pm$0.068 \\
PSVAE & 0.854$\pm$0.003 & 0.836$\pm$0.004 & 0.839$\pm$0.005 \\
PSVAE-SA & 0.847$\pm$0.003 & 0.816$\pm$0.008 & \textbf{0.832}$\pm$0.002 \\
\method & \textbf{0.837}$\pm$0.001 & \textbf{0.813}$\pm$0.006 & 0.833$\pm$0.004 \\
\bottomrule
\end{tabular}
\end{table}

\subsection{Main Results}
Table \ref{tab:coverage} and \ref{tab:cost} show the coverage and cost on the test set of three datasets with $k=10$ after 20 iterations. We ran the experiments 5 times with different random seeds and calculated the average and standard deviations. 
Our method, \method, performs better than the best baseline model, PSVAE-SA, with a gain of 8\% increase in coverage and a cost reduction of 1.23\% on AIDS. It also performs best on Mutagenicity in terms of coverage and cost. Dipole has the highest coverage with a similar cost.
It means that the explanations from \method are similar to the input molecules with different GNN predictions. 
This helps people understand the GNN predictor's behaviors on the whole input dataset with limited molecules.

Figure \ref{fig:topk} shows how the candidate set size $k$ impacts performance. \method has the highest coverage with different $k$. When $k$ increases, the performance gap grows. All methods reach their maximum coverage on the input graph set after $k=25$. The candidate set covers almost the entire input graph set. 
Two generative baselines, PSVAE and PSVAE-SA, do better than GCFExplainer when $k$ is small. But GCFExplainer reduces the performance gap as the number of candidates grows.

In general, \method outperforms other baselines in cost. On AIDS and Dipole, GCFExplainer can match the input set at a lower cost than PSVAE and PSVAE-SA with $k=50$. This is similar to \method. However, on Mutagenicity, our method \method has a clear edge over the other baselines for all $k$.
PSVAE-SA has a small cost at $k=1$ on three datasets while it lags when $k$ grows larger. This means that the simulated annealing starting from individual input is good at finding the local optima explanation but not at the global optima. However, \method can find a better global explanation set closer to the input set for different $k$.

\subsection{Analysis} 
\label{sec:analysis}

\begin{table}[!htp]
\centering
\caption{Ablation Study on \method. Cov. inidated Coverage. All experiments are based on $i=20$, $k=10$ and beam size $10$.}
\label{tab:ablation}\small\setlength{\tabcolsep}{3pt}
\begin{tabular}{lcccccc}\toprule
&\multicolumn{2}{c}{AIDS} &\multicolumn{2}{c}{Mutagenicity} &\multicolumn{2}{c}{Dipole} \\
 & Cov. $\uparrow$ & Cost $\downarrow$ & Cov. $\uparrow$ & Cost $\downarrow$ & Cov. $\uparrow$ & Cost $\downarrow$ \\
\midrule
\method &0.811 &0.836 &0.914 &0.813 & 0.887 &0.831 \\
\midrule
w/o PSVAE &0.168 &0.923 & 0.900 & 0.799 &0.713 &0.846 \\
w/o Adapter &0.615 &0.857 & 0.887 & 0.791 &0.817 &0.833 \\
\bottomrule
\end{tabular}
\end{table}

\textbf{Ablation Studies}
In Table \ref{tab:ablation}, we further investigate the performance of our model with several ablation studies. These studies delve deeper into the fundamental components of the \method model and their contributions to the model's overall effectiveness.
To better understand the role of the PSVAE in \method, we consider a variant of \method, namely \method w/o PSVAE. This variant starts with randomly initialized parameters and is trained from scratch. Noticeably, the absence of the pre-trained PSVAE results in a significant reduction in coverage on the AIDS and Dipole datasets. Nonetheless, it is still able to generate valid explanations because of the subgraph-based generation method.
We also study the influence of the trained adapter by examining a version of \method (referred to as \method w/o trained Adaptor) that includes a randomly initialized adapter without any further training. The adapter in this context is directly used for inference with a random shift on the latent space. The results highlight that the removal of the adapter leads to decreases in coverage and an increase in cost. This is primarily due to the alignment mismatch with the human-designed principles.

\begin{figure}
    \centering
    \includegraphics[width=0.9\linewidth]{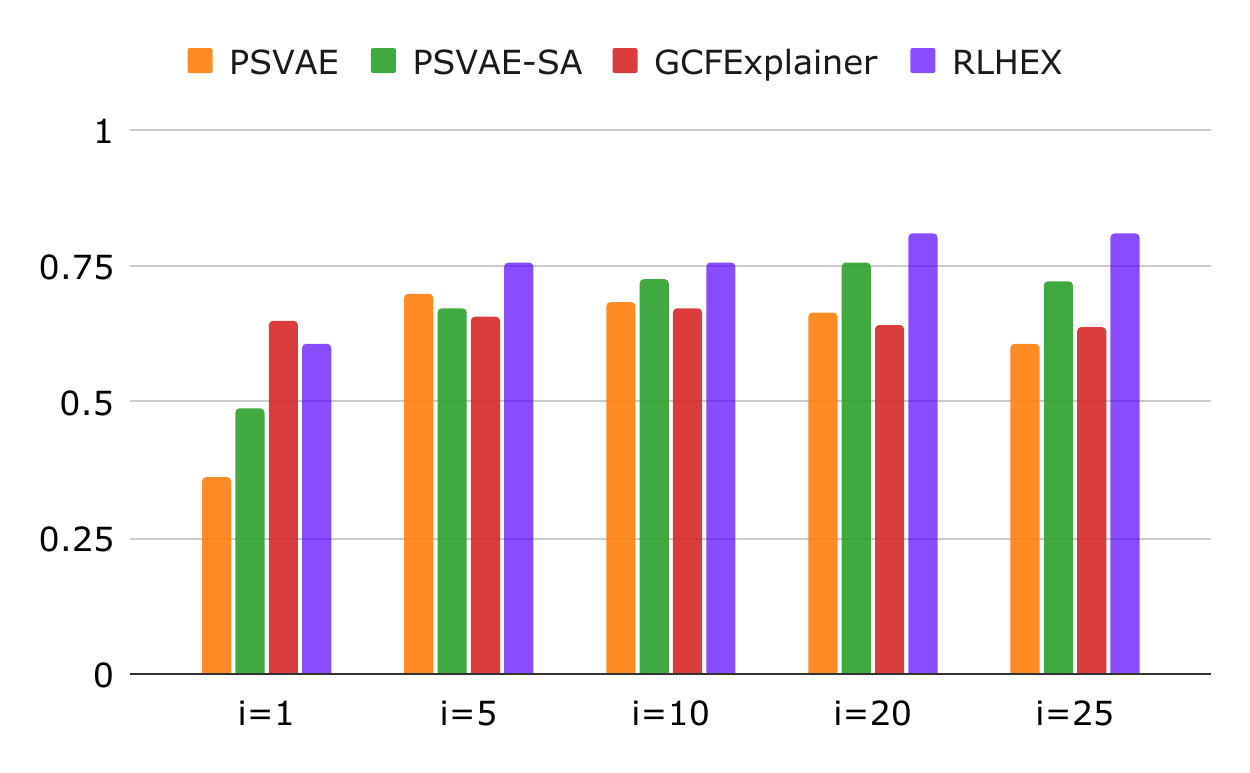}
    \caption{Coverage on AIDS dataset with different iterations $i$. We limit the explanation set with $k=10$ to calculate the coverage.}
    \label{fig:coverage_iter}
\end{figure}

\textbf{\method achieves the highest coverage after one iteration}
Figure \ref{fig:coverage_iter} illustrates how the performance is impacted by the number of iterations during the inference process. For the PSVAE-based model, an iteration is defined as one pass over the set of input graphs, $\mathbb{G}$. For GCFExplainer, the maximum step of the random walk is defined as $i * |\mathbb{G}|$, where $i$ denotes the iteration number.
As shown in Figure \ref{fig:coverage_iter}, \method reaches optimal performance after one iteration and maintains a stable performance after 20 iterations. However, GCFExplainer shows the best performance at the first iteration. As the iteration number increases, other models start to outperform GCFExplainer. 
This performance decrease can be attributed to GCFExplainer's process of exploring every possible perturbation on nodes and edges for the current molecule at each step, which includes investigating up to 100,000 neighbors. In comparison, PSVAE-based methods explore 10 candidates (beam size = 10) for each input graph.
Consequently, GCFExplainer's large search space at each step makes it more effective when the iteration number is small. However, as the number of iterations increases, its performance deteriorates due to its inefficient exploration strategy.

\begin{figure*}[htb]
    \centering
    \includegraphics[width=0.9\linewidth]{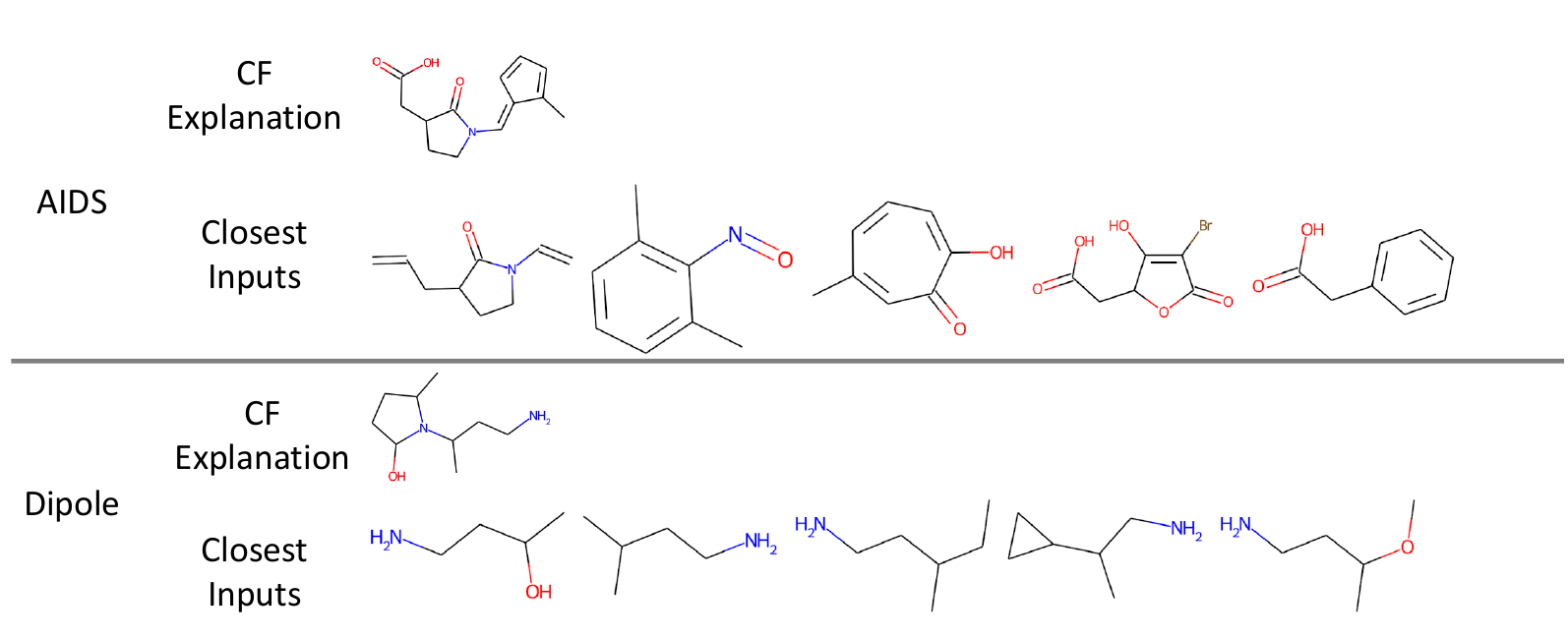}
    \caption{The counterfactual (CF) explanation generated for the closest input molecules from the AIDS and Dipole datasets. For each CF explanation, we compute the distance between it and the input molecules, selecting the top 5 input molecules for display. The generated CF explanation for AIDS exhibits a coverage of 0.231 over the input molecule set, while the CF explanation for the Dipole dataset shows a coverage of 0.209.}
    \label{fig:case}
\end{figure*}

\textbf{Generated CF explanations are more interpretable}
Figure \ref{fig:case} displays two cases produced by \method, specifically focusing on the AIDS and Dipole datasets. In the AIDS dataset, the GNN predictor classifies the input graphs as negative, or label=0, indicating that these molecules are inactive against AIDS. Conversely, in the Dipole dataset, the input molecules are deemed non-polar, while the CF explanation shows them as polar.
As seen in Figure \ref{fig:case}, the CF explanation closely resembles these input molecules, as they share several common chemical sub-graphs. This similarity allows \method to encapsulate the behavior of the GNN predictor across various input molecules by grouping multiple negative molecules together. It further uncovers the necessary conditions that would prompt the GNN predictor to alter its prediction.
Another highlight of our work is that by comparing the varying sub-graphs between the CF explanation and the covered input molecules, domain experts can more easily identify which functional group the GNN predictor uses to make its decisions. This feature makes \method a valuable tool for enhancing understanding and facilitating decision-making in chemists' work.

\subsection{Expert Assessment on CF Explanation}
We also engaged molecular chemists to evaluate our CF explanations. Although the evaluations lacked empirical laboratory testing, the expert feedback was in general alignment with the explanations about specific classes of molecules. Through this procedure, the chemists could assess the efficacy of the GNN classifier through its alignment with known chemical knowledge. Based on the case shown in Figure \ref{fig:case}, chemists made the following observations:

\textbf{AIDS}
The CF candidate is predicted to be active against AIDS. Although the input negative molecules also have the fused aromatic rings and hydroxyl (-OH) groups that could potentially form important interactions with viral targets, the fused 3-ring system of the CF candidate increases its possibility of being against AIDS. The fused 3-ring system resembles known HIV integrase inhibitor pharmacophores~\cite{Das:2005aa}.

\textbf{Dipole}
The CF candidate and the covered input molecules have O-H and N-H bonds, which are polar due to the electronegativity difference between oxygen and hydrogen\cite{PhysRevLett.108.058301}. However, the CF candidate has a bent geometry, which allows the bond dipoles to add up and create a net molecular dipole~\cite{Griffiths_Schroeter_2018}.

\section{Conclusion and Discussion}
\label{sec:conclusion}

In this paper, we introduced \method, a global counterfactual explanation method that strives to aid domain experts to better understand GNN predictions. Our proposed model aligns with the domain experts' criteria and uses Proximal Policy Optimization (PPO) to generate chemically valid explanations that can cover the highest number of input molecules. Importantly, our method takes into account the interpretability requirement of domain experts, an aspect often overlooked in CF explanations, making it suitable for understanding molecular property predictions.
Experimental results show that \method outperforms other strong baselines on three real-world molecular datasets. 

Further work can focus on the scalability of our method. We plan to investigate \method's performance on larger datasets and more complex molecular structures. Although the primary application of our method is in cheminformatics, we anticipate that the underlying principle of \method could be broadly applied across various fields requiring interpretability in the use of GNNs. Furthermore, more human-designed principles can be flexibly integrated into \method to align different preferences of domain experts.

To conclude, \method represents a significant step towards creating more interpretable and understandable GNN models. By generating global counterfactual explanations through human-aligned principles, \method offers a promising avenue for domain experts to better utilize and comprehend the findings of GNN predictions, particularly in the evolving generative landscape. Through our work, we aspire to bridge the gap between the scale and complexity of sequence-based prediction models and the intuitiveness required by experts to make new scientific discoveries.

\section*{Acknowledgement}
\label{sec:acknowledge}
We gratefully acknowledge the support of the National Science Foundation (\# 2229876) for funding this research.

\newpage
\bibliographystyle{ACM-Reference-Format}
\balance
\bibliography{reference}

\end{document}